\documentclass[final]{cvpr}

\usepackage{times}
\usepackage{epsfig}
\usepackage{graphicx}
\usepackage{amsmath}
\usepackage{amssymb}

\usepackage[pagebackref=true,breaklinks=true,colorlinks,bookmarks=false]{hyperref}

\usepackage[utf8]{inputenc} %
\usepackage[T1]{fontenc}    %
\usepackage{url}            %
\usepackage{booktabs}       %
\usepackage{amsfonts}       %
\usepackage{nicefrac}       %
\usepackage{microtype}      %
\usepackage{natbib}
\setcitestyle{numbers,square}
\renewcommand{\cite}{\citep}

\usepackage[subrefformat=parens,labelformat=parens]{subfig}
\graphicspath{{figs/}}
\DeclareGraphicsExtensions{.pdf,.jpg,.png,.eps}

\usepackage[ruled,linesnumbered]{algorithm2e}
\usepackage{xspace}
\usepackage{booktabs}
\usepackage{amsthm}
\usepackage[toc,page,title,titletoc]{appendix}
\usepackage{wrapfig}  %
\usepackage{makecell}  %
\usepackage{siunitx} %
\usepackage{multirow}

\usepackage{floatrow}

\usepackage{enumitem}
\setitemize{noitemsep,topsep=0pt,parsep=0pt,partopsep=0pt}

\def\abovestrut#1{\rule[0in]{0in}{#1}\ignorespaces}

\newcommand{\smallgap}{\abovestrut{0.20in}}

\DeclareRobustCommand*\pct{\scalebox{.9}{\%}}

\newcommand{\myparagraph}[1]{\textbf{#1}.~} %
\newcommand{\mytimes}{\medmuskip=0mu\times}
\newcommand{\dense}{\thickmuskip=2mu}

\newcommand{\obs}{\ensuremath{o}}
\newcommand{\state}{\ensuremath{s}}

\newcommand{\localmap}{\ensuremath{m}}
\newcommand{\globalmap}{\ensuremath{M}}
\newcommand{\mapchannels}{\ensuremath{N_\mathrm{ch}}}

\def\cvprPaperID{7658} %
\def\confYear{CVPR 2021}

\begin{document}

\title{Differentiable SLAM-net: Learning Particle SLAM for Visual Navigation}

\author{Peter Karkus \quad Shaojun Cai \quad David Hsu\\
National University of Singapore\\
{\tt\small karkus@comp.nus.edu.sg}
}

\maketitle

\begin{abstract}
  Simultaneous localization and mapping~(SLAM) remains challenging for a number of downstream applications, such as visual robot navigation, because of rapid turns, featureless walls, and poor camera quality.
  We introduce the Differentiable SLAM Network (SLAM-net) along with a navigation architecture to enable planar robot navigation in previously unseen indoor environments. SLAM-net encodes a particle filter based SLAM algorithm in a differentiable computation graph, and learns task-oriented neural network components by backpropagating through the SLAM algorithm. 
  Because it can optimize all model components jointly for the end-objective, SLAM-net learns to be robust in challenging conditions. 
  We run experiments in the Habitat platform with different real-world RGB and RGB-D datasets. SLAM-net significantly outperforms the widely adapted ORB-SLAM in noisy conditions. Our navigation architecture with SLAM-net improves the state-of-the-art for the Habitat Challenge 2020 PointNav task by a large margin (37\pct{} to 64\pct{} success). Project website: \url{http://sites.google.com/view/slamnet}
  
\end{abstract}

\section{Introduction}

\begin{figure}[!t]
\centering
\includegraphics[width=0.95\textwidth]{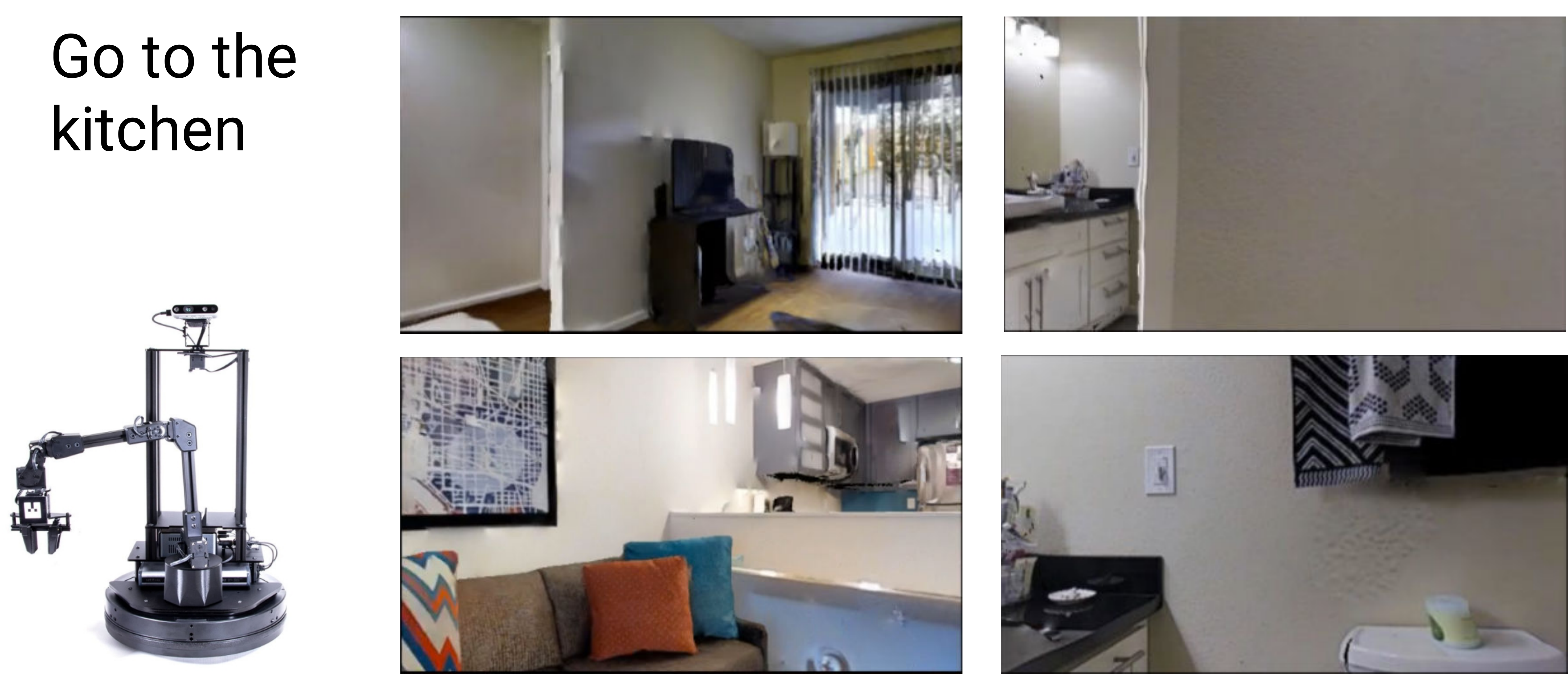}
\caption{Visual robot navigation is challenging for SLAM, \eg, because the robot frequently faces featureless walls; it rotates quickly; the onboard camera produces noisy images; the frame rate is low; \etc{} The images were take by our navigation agent in the Habitat environment.
}
\label{fig:challenges}
\end{figure}

Modern visual SLAM methods achieve remarkable performance when evaluated on suitable high-quality data~\cite{fuentes2015visual}. However, in the context of downstream tasks, such as indoor robot navigation, a number of difficulties remain~\citep{cadena2016past, mishkin2019benchmarking, chen2020survey}. An imperfect navigation agent captures substantially different images from a human, as it may frequently face featureless walls and produce rapid turns~(see \figref{fig:challenges}). Further, despite advances in sensor technology, modern robots still often use cameras with noisy images, low frame rate, and a narrow field-of-view~\cite{habitatchallenge}. These factors make  feature extraction and association difficult. Relocalization and loop closure can be challenging due to environmental changes and repetitive feature.
Finally, integrating SLAM into a navigation pipeline is not trivial, because the map representation must be suitable for downstream planning, it may need to capture task-dependent information, and planning must be able to handle map imperfections.

This paper introduces the Differentiable SLAM Network (SLAM-net) together with a navigation architecture for downstream indoor navigation. The key idea of SLAM-net is to encode a SLAM algorithm in a differentiable computation graph, and learn neural network model components for the SLAM algorithm end-to-end, by backpropagating gradients through the algorithm. Concretely SLAM-net encodes 
the particle filter based FastSLAM algorithm~\cite{montemerlo2002fastslam} and learns mapping, transition and observation models. SLAM-net fills a gap in the literature on differentiable robot algorithms~\cite{tamar2016value, gupta2017cognitive, karkus2018particle, karkus2019differentiable}.

The benefit of SLAM-net compared to unstructured learning approaches is that its encoded particle filter provides a strong prior for learning. The benefit over classic SLAM is that all components are learned, and they are directly optimized for the end-objective. Concretely, SLAM-net learns RGB and RGB-D observation models for the encoded FastSLAM algorithm, which previously relied on handcrafted models and lidar sensors. Further, because of the task-oriented learning, feature extractors can learn be more robust against domain specific challenges, \eg, faced with downstream navigation; while on the flip side they may be less reusable across tasks.

\begin{figure*}[!t]
\centering
\includegraphics[width=0.9\textwidth]{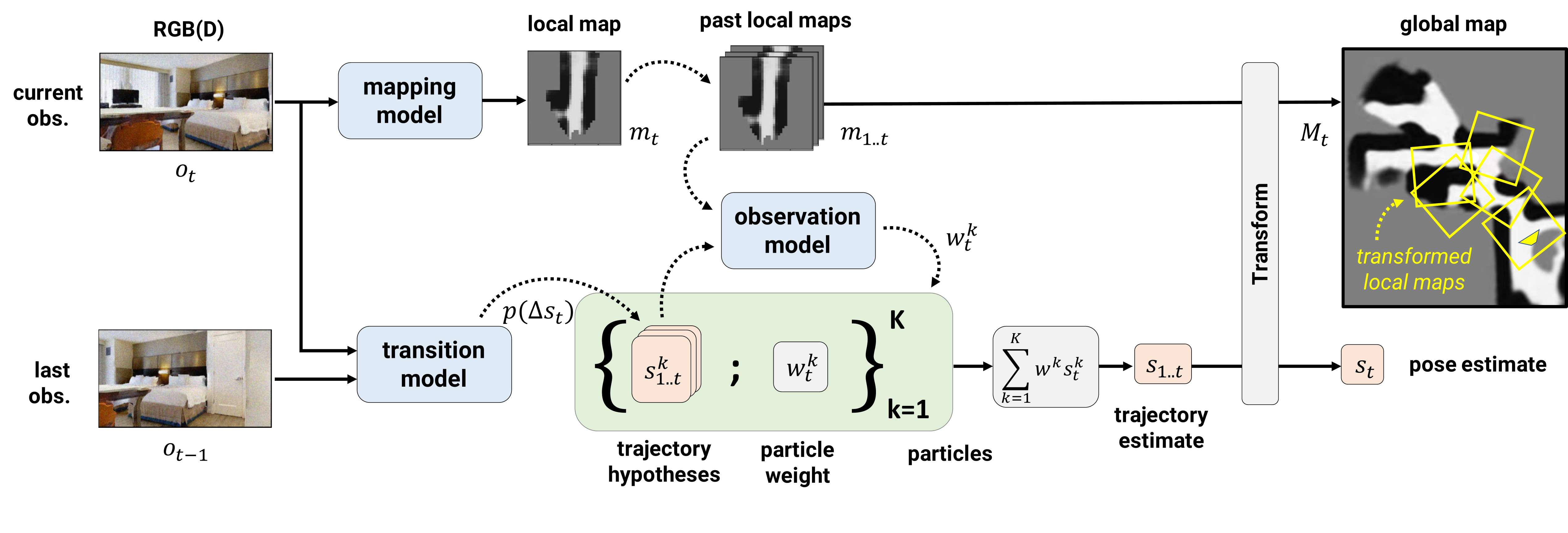}
\vspace{-0.2cm}  %
\caption{Differentiable SLAM-net. The global map is maintained by a collection of learned local grid maps. The trajectory is tracked by a particle filter. Particles represent trajectories and they are updated with learned neural network components: the mapping, transition, and observation models.}
\label{fig:slamnet}
\end{figure*}

We validate SLAM-net for localization with RGB and RGB-D input, as well as  downstream robot navigation in previously unseen indoor environments. We use the Habitat simulation platform~\citep{savva2019habitat} with three  real-world indoor scene datasets. We additionally experiment with the KITTI visual odometry data~\cite{geiger2012we}.  SLAM-net achieves good performance under challenging conditions where the widely used ORB-SLAM~\citep{mur2017orb} completely fails; and trained SLAM-nets transfer over datasets. For downstream navigation we propose an architecture similar to Neural SLAM~\cite{chaplot2020learning}, but with using our differentiable SLAM-net module. Our approach significantly improves the state-of-the-art for the CVPR Habitat 2020 PointNav challenge~\cite{habitat2020leaderboard}. %

\section{Related work}

\myparagraph{Learning based SLAM}
Learning based approaches to SLAM have a large and growing literature. For example, CodeSLAM\citep{bloesch2018codeslam} and SceneCode\citep{zhi2019scenecode} learn a compact representation of the scene; CNN-SLAM\citep{tateno2017cnn} learns a CNN-based depth predictor as the front-end of a monocular SLAM system;  BA-net \cite{tang2018ba} learns the feature metric representation and a damping factor for bundle adjustment.
While these works use learning, they typically only learn specific modules in the SLAM system. Other approaches do end-to-end learning but they are limited to visual odometry, \ie, they estimate relative motion between consecutive frames without a global map representation~\cite{zhou2017unsupervised, li2018undeepvo}. Our method maintains a full SLAM algorithm and learns all of its components end-to-end.

\myparagraph{Classic SLAM}
Classic SLAM algorithms can be divided into filtering and optimization based approaches \cite{strasdat2012visual}. Filtering-based approaches maintain a probability distribution over the robot trajectory and sequentially update the distribution with sensor observations \cite{civera20091,azarbayejani1995recursive,davison2003real,montemerlo2002fastslam,montemerlo2003fastslam}. 
Optimization-based approaches apply bundle adjustment on a set of keyframes and local maps;   %
and they are popular for both visual \cite{strasdat2012visual, mur2015orb,mur2017orb,klein2007parallel} and lidar-based SLAM~\cite{hess2016real}.
Our approach builds on a filtering-based algorithm, FastSLAM~\cite{montemerlo2002fastslam, montemerlo2003fastslam}.
The original algorithm (apart from a few adaptations~\cite{barfoot2005online,  lee2011rs, hartmann2012real}) works with a lidar sensor and hand-designed model components. Robot odometry information is typically used for its transition model, and either landmarks~\cite{montemerlo2002fastslam} or occupancy grid maps~\cite{grisetti2007improved} are used for its observation model. In contrast, we learn neural network models for visual input by backpropagation through a differentiable variant of the algorithm. We choose this algorithm over an optimization based method because of the availability of differentiable particle filters~\cite{jonschkowski2018differentiable, karkus2018particle}, and the suitability of the algorithm for downstream robot navigation.

\myparagraph{Differentiable algorithms}
Differentiable algorithms are emerging for a wide range of learning domains, including state estimation~\citep{haarnoja2016backprop, jonschkowski2018differentiable, karkus2018particle, ma2020particle}, visual mapping~\citep{gupta2017cognitive, karkus2020differentiable}, planning~\citep{tamar2016value, karkus2017qmdp, farquhar2017treeqn,  oh2017value, guez2018learning, yonetani2020path} and control tasks~\citep{amos2018differentiable, okada2017path, east2019infinite, bhardwaj2020differentiable}. Differentiable algorithm modules have been also composed together for visual robot navigation~\cite{gupta2017cognitive, karkus2019differentiable, ma2020discriminative}. This work introduces a differentiable SLAM approach that fills a gap in this literature.
While \citet{jatavallabhula2019gradslam} have investigated differentiable SLAM pipelines, they focus solely on the effect of differentiable approximations and do not perform learning of any kind.

\myparagraph{Visual navigation}
A number of learning based approaches has been proposed for visual navigation recently~\citep{gupta2017cognitive, anderson2018evaluation,  mishkin2019benchmarking, karkus2019differentiable, wijmans2019dd, chaplot2020learning, chaplot2020neural, ramakrishnan2020occupancy}. 
Modular approaches include CMP~\cite{gupta2017cognitive}, DAN~\cite{karkus2019differentiable} and Neural SLAM~\cite{chaplot2020learning}. However, CMP assumes a known robot location, circumventing the issue of localization. DAN assumes a known map that is given to the agent.  Neural SLAM~\cite{chaplot2020learning, ramakrishnan2020occupancy} addresses the joint SLAM problem, but it relies solely on relative visual odometry without local bundle adjustment or loop closure, and thus it inherently accumulates errors over time.
We propose a similar navigation architecture to Neural SLAM~\cite{chaplot2020learning}, but utilizing our Differentiable SLAM-net module in place of learned visual odometry.  %

\section{Differentiable SLAM-net}\label{sec:slamnet}

\subsection{Overview}
The Differentiable SLAM-net architecture is shown in~\figref{fig:slamnet}. Inputs are RGB(D) observations $\obs_t$, outputs are pose estimate $\state_t$ and global map $\globalmap_t$. SLAM-net assumes the robot motion is (mostly) planar. Poses are 2D coordinates with 1D orientation; the global map is a 2D occupancy grid.

Internally SLAM-net represents the global map as a collection of local maps, each associated with a local-to-global transformation. Local maps are $\dense N\mytimes N \mytimes \mapchannels$ grids that, depending on the configuration, may encode occupancy and/or learned latent features. 
We add a local map for each observation, but without knowing the robot pose we do not know the correct local-to-global map transformation. Instead, the algorithm maintains a distribution over the unknown robot trajectory and closes loops using particle filtering~\cite{doucet2001introduction}. 
Our algorithm is based on FastSLAM~\cite{montemerlo2002fastslam,montemerlo2003fastslam}, and our differentiable implementation is built on PF-nets~\cite{karkus2018particle}. 

The algorithm works as follows. The particle filter maintains $K$ weighted particles, where each particle represents a trajectory $\state_{0:t}^k$. At $t=0$ all particle trajectories are set to the origin; particle weights are constant, and the local map collection is empty. In each time step a \emph{mapping model} predicts a local map $\localmap_t$ from the input observation $\obs_t$, and $\localmap_t$ is added to the collection. Particle trajectories are extended with samples from a probabilistic \emph{transition model} that estimates the relative motion given $\obs_t$ and $\obs_{t-1}$. Particle weights are then updated using an \emph{observation model} which measures the compatibility of $\localmap_t$ and the past local maps $\localmap_{1:t-1}$ assuming the particle trajectory $\state_{0:t}^k$ was correct. The pose output is obtained by taking the weighted sum of particle trajectories. Optionally, a global occupancy grid map is obtained with simple 2D image transformations that combine local maps along the mean trajectory.

The key feature of SLAM-net is that it is end-to-end differentiable. That is, the mapping, transition and observation models are neural networks, and they can be trained together for the end-objective of localization accuracy (and/or global map quality). To make the algorithm differentiable we use the reparameterization trick~\citep{kingma2013auto} to differentiate through samples from the transition model; and we use spatial transformers~\cite{jaderberg2015spatial} for differentiable map transformations. The rest of the operations of the algorithm, as presented, are already differentiable. While not used in our experiments, differentiable particle resampling could be incorporated from prior work~\citep{karkus2018particle, zhu2020towards, corenflos2021differentiable}.
Further, due to limited GPU memory, to make use of the differentiable algorithm for learning our design choices on the local map representation and the formulation of the observation model are important.

Next we introduce each component of SLAM-net. Network architecture details are in the Appendix.

\begin{figure*}[t] %
\centering
\includegraphics[width=0.9\textwidth]{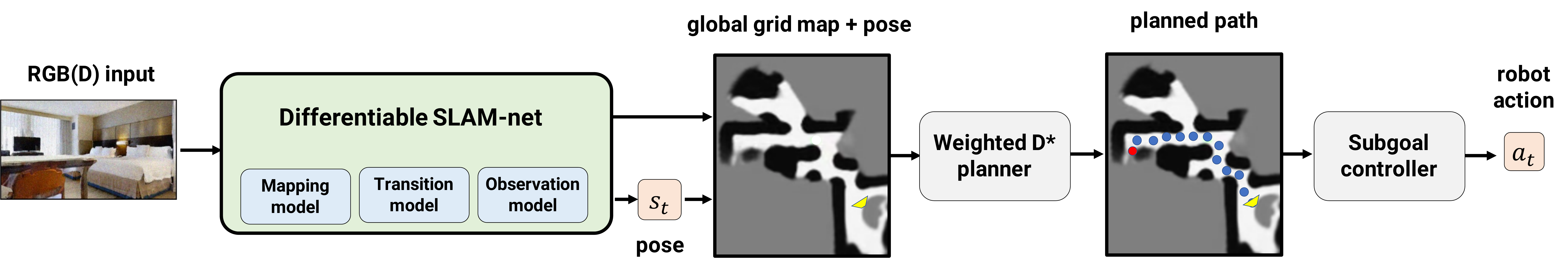}
\caption{Visual navigation pipeline with the Differentiable SLAM-net, a path planner, and a motion controller.} %
\label{fig:navpipeline}
\end{figure*}

\subsection{Transition model}
The transition model is a CNN that takes in the concatenated current and last  observations, $\obs_t$ and $\obs_{t-1}$, and outputs parameters of Gaussian mixture models with separate learned mean and variance for the relative 2D pose and 1D orientation. The transition model is pre-trained to maximize the log-likelihood of true relative poses along the training trajectories. It is then finetuned together with the rest of the SLAM-net components optimizing for the end-objective.

\subsection{Mapping model}
The mapping model is a CNN with a pre-input perspective transformation layer. The input is observation $o_t$, the output is local map $m_t$. Local maps serve two purposes: to be fed to the observation model and aid pose estimation by closing loops; and to construct a global map for navigation. 

Our local maps are $\dense N\mytimes N \mytimes \mapchannels$ grids that capture information about the area in front of the robot. In one configuration local maps encode occupancy, \ie, the probability of the area being occupied for the purpose of navigation. This model is trained with a (per cell) classification loss using ground-truth occupancy maps. In another configuration local maps encode learned latent features that have no associated meaning. This model is trained for the end-objective by backpropagating through the observation model. 
In both cases we found it useful to add an extra channel that encodes the visibility of the captured area. For depth input this is computed by a projection; for RGB input it is predicted by the network, using projected depth for supervision.

\subsection{Observation model}
The observation model is the most important component of the architecture. It updates particle weights based on how ``compatible'' the current local map $\localmap_t$ would be with past local maps $\localmap_{1:t-1}$ if the particle trajectory $\state_{0:t}^k$ was correct. Intuitively we need to measure whether local maps capture the same area in a consistent manner. Formally we aim to estimate a compatibility value proportional to the log-probability $\log{p(\localmap_t | \localmap_{1:t-1}, \state_{1:t}^k)}$. 

We propose a discriminative observation model that compares pairs of local maps with a learned CNN. The CNN takes in a current local map $\localmap_t$ and a past local map $\localmap_\tau$ concatenated, and outputs a compatibility value. Importantly, the past local map is transformed to the viewpoint of the current local map according to the relative pose in the particle trajectory $(s_t^k, s_\tau^k)$. We use spatial transformers~\cite{jaderberg2015spatial} for this transformation.
The overall compatibility is the sum of pairwise compatibly values along the particle trajectory. Compatibility values are estimated for all particles. Particle weights are then updated by multiplying with the exponentiated compatibility values, and they are normalized across particles. CNN weights are shared.

For computational reasons, instead of comparing all local map pairs, we only compare the \emph{most relevant} pairs. During training we pick the last 4--8 steps; during inference we dynamically choose 8 steps based on the largest overlapping view area (estimated using simple geometry).

\subsection{Training and inference}
Our training data consists of image-pose pair trajectories (depth or RGB images, and 2D poses with 1D orientation); and optionally ground-truth global occupancy maps for pre-training the mapping model. The end-to-end training objective is the sum of Huber losses for the 2D pose error and 1D orientation error. 

We train in multiple stages.  We first pre-train the transition model. We separately pre-train the observation model together with the mapping model for the end-objective, but in a low noise setting. That is, in place of the transition model we use ground truth relative motion with small additive Gaussian noise.  Finally we combine all models and finetune them together for the end-objective. During finetuning we freeze the convolution layers and mixture head of the transition model. When the mapping model is configured to predict occupancy it is  pre-trained separately and it is frozen during finetuning.
 
An important challenge with training SLAM-net is the computational and space complexity of backpropagation through a large computation graph. To overcome this issue, during training we use only short trajectories (4-8 steps), and $\dense K=32$ particles without resampling. During inference we use the full length trajectories, and by default $\dense K=128$ particles that are resampled in every step.

\subsection{Implementation details}
We implement SLAM-net in Tensorflow~\cite{tensorflow2015-whitepaper} based on the open-source code of PF-net~\cite{karkus2018particle}. We adopt the training strategy where the learning rate is decayed if the validation error does not improve for 4 epochs. We perform 4 such decay steps, after which training terminates, and the model with the lowest validation error is stored.  The batch size is 16 for end-to-end training, and 64 for pre-training the mapping and transition models. We use Nvidia GeForce GTX 1080 GPUs for both training and inference.

For RGB input we configure local maps with 16 latent feature channels that are not trained to predict occupancy. For RGB-D input local maps are configured with both latent channels and one channel that predicts occupancy. Further, with RGB-D data we only use depth as input to SLAM-net.

\section{Visual Navigation with SLAM-net}\label{sec:navigation}
We propose a visual navigation pipeline (\figref{fig:navpipeline}) that combines SLAM-net with modules for path planning and motion control.
In the pipeline SLAM-net periodically predicts the robot pose and a global occupancy grid map. %
The map and pose are fed to a 2D path planner that plans a path to the goal. The path is then tracked by a local controller the outputs robot actions. %

\myparagraph{Task specification}
We follow the task specification of the Habitat 2020 PointNav Challenge~\cite{habitatchallenge}. A robot navigates to goals in previously unseen apartments using noisy RGB(D) input. The goal is defined by coordinates relative to the initial pose, but the robot location is unknown thereafter, and discrete robot actions generate noisy motion. %
Navigation is successful if the robot takes a dedicated \emph{stop} action within 0.36 meters to the goal. %
Note that this success criteria places high importance on pose estimation accuracy.

\myparagraph{Path planner}
The challenge of path planning in the context of visual navigation is the imperfect partial knowledge of the map and the robot pose. To address this we adopt a weighted variant of the D* algorithm~\citep{koenig2005fast} with costs that penalize moving near obstacles. 
In each step when the map and pose are updated by the SLAM-net, the path is replanned. For planning we convert the occupancy grid map to an 8-connected grid where cells are assigned a cost. 
We threshold the map ($\dense p>0.5$ is an obstacle, $\dense p<=0.5$ is free space) and define cell costs based on the distance to the nearest obstacle. 

Additionally, we use a collision recovery mechanism. Upon detecting a collision an obstacle is registered to the map at the current estimated pose. The robot is then commanded to turn around (6 turn actions) and takes a step back (1 forward action). Collisions are not directly observed. We trigger collision recovery if the estimated pose does not change more than $3cm$ following a forward action. A similar mechanism was proposed in \citet{chaplot2020learning}.

We also hard-coded an initial policy that makes the robot turn around in the beginning of an episode. The initial policy terminates once the estimated rotation exceeds $370^{\circ}$.

\myparagraph{Local controller}
The planned path is tracked by a simple controller that chooses to turn or move forward aiming for the furthest straight-line traversable point along the path. The output of the controller are discrete actions.

\section{Experiments}\label{sec:experiments}
Our experiments focus on the following questions. 1)~Can SLAM-net learn localization in previously unseen indoor scenes? How does it compare to existing methods? 2)~Does SLAM-net enable downstream robot navigation? 3)~What does SLAM-net learn and how do model components and hyper parameters affect performance? 4)~Do learned models transfer to new datasets? 5)~What are the limitations of SLAM-net if applied, \eg, to  autonomous driving data?

\subsection{Datasets}

\myparagraph{Habitat} Our main experimental platform is the Habitat simulator~\cite{savva2019habitat} configured with different real-world datasets: Gibson~\cite{xiazamirhe2018gibsonenv},  Replica~\cite{replica19arxiv}, and Matterport~\cite{Matterport3D}. The datasets contain a large number of 3D scans of real world indoor scenes, typically apartments.  Habitat embeds the scenes in an interactive physics simulator for robot navigation. The simulator renders photorealistic but noisy first-person RGB(D) observations and simulates realistic robot motion dynamics. The camera has a horizontal FOV of $70^{\circ}$ and a resolution of $640\mytimes360$. For SLAM-net we downscale images to $160\mytimes90$. Depth values are in the range of 0.1 to 10 meters. 
Unless stated otherwise, we use the Habitat Challenge 2020 configuration~\cite{habitatchallenge}: Gaussian noise for RGB images; the Redwood noise model for depth images~\cite{choi2015robust}; the Locobot motion model for actuation noise~\cite{murali2019pyrobot}. To train and evaluate SLAM methods we generate a fixed set of trajectories. For navigation we let our method interact with the simulator.

\myparagraph{Gibson data}\label{sec:datasets}
Following \citet{savva2019habitat} we use 72 scenes from the Gibson dataset for training and further split the original validation set to 7 scenes for validation and 7 scenes for testing. 
We use 36k of the provided navigation episodes for training (500 per scene). Given a start and goal pose we generate a trajectory with a navigation policy that switches between a shortest-path expert (30 steps) and random actions (40 steps).
For evaluation we generate 105 trajectories (15 per test scene) using three different navigation policies: the shortest-path expert 
(\textbf{traj\_expert}); the shortest-path expert mixed with random actions~(\textbf{traj\_exp\_rand}); and our final navigation pipeline~(\textbf{traj\_nav}).

\myparagraph{Replica and Matterport data}
We use the Matterport and Replica datasets for transfer experiments without additional training. We generate trajectories for evaluation similarly to the Gibson data, using the shortest-path expert policy. We use the original validation split with a total of 170 and 210 trajectories for Replica and Matterport respectively.

\myparagraph{KITTI data}
We conduct additional experiments with the KITTI real world driving data~\cite{geiger2012we}. We use the full KITTI raw dataset for training, validation, and testing. Following the KITTI Odometry Split~\cite{geiger2012we} the validation trajectories are 06 and 07, and the testing trajectories are 09 and 10. Since the original depth information for KITTI dataset is from sparse lidar, we use the completed depth data from the \cite{uhrig2017sparsity} as ground-truth depth.

\myparagraph{Statistics}
\tabref{tab:datastat} provides statistics for each set of evaluation trajectories. We provide the mean and standard deviation of the trajectory length, number of frames, and number of turn actions (where applicable).

\begin{table}[t!]
\centering
\scalebox{0.85}{
\begin{tabular}{lccc}
\toprule
Dataset  & length[m]  & \#frames  & \#turns \\
 \midrule
Gibson ({traj\_expert})       &  7.4{\small$\pm$3.8}   &  51.1{\small$\pm$24.7}   &  22.6{\small$\pm$11.6}   \\
Gibson ({traj\_exp\_rand})  & 14.5{\small$\pm$7.2}   &  152.3{\small$\pm$67.9}   &  75.0{\small$\pm$34.6}\\
Gibson ({traj\_nav})  & 11.9{\small$\pm$8.8}   &  117.0{\small$\pm$113.3}   &  74.1{\small$\pm$84.0} \\
Matterport ({traj\_expert}) & 13.0{\small$\pm$7.0}   &  82.0{\small$\pm$37.9}   &  32.9{\small$\pm$14.1} \\
Replica ({traj\_expert}) & 8.0{\small$\pm$2.8}   &  53.9{\small$\pm$17.7}   &  23.5{\small$\pm$9.17} \\   
KITTI-09  & 1680.3 & 1551 \\ 
KITTI-10  & 910.48  & 1161 \\ 
 \bottomrule
\end{tabular}
}%
{\vspace{0cm}\caption{Data statistics.}\label{tab:datastat}}
\end{table}

\begin{table*}
\centering
\scalebox{0.9}{
\begin{tabular}{l@{\hskip 0.8cm}c@{\hskip 1.0cm}cc@{\hskip .4cm}cc@{\hskip .4cm}cc@{\hskip 1.0cm}cc}
\toprule
Sensor  & \textbf{}
  & \multicolumn{2}{c@{\hskip .4cm}}{\textbf{RGBD}} 
  & \multicolumn{2}{c@{\hskip .4cm}}{\textbf{RGBD}} 
  & \multicolumn{2}{c@{\hskip 1.0cm}}{\textbf{RGBD}}  & \multicolumn{2}{c}{\textbf{RGB}} \\    
Trajectory generator  & \textbf{}
  & \multicolumn{2}{c@{\hskip .4cm}}{\textbf{traj\_expert}} 
  & \multicolumn{2}{c@{\hskip .4cm}}{\textbf{traj\_exp\_rand}} 
  & \multicolumn{2}{c@{\hskip 1.0cm}}{\textbf{traj\_nav}}  
  & \multicolumn{2}{c}{\textbf{traj\_expert}}  \\    
Metric  &  {\small runtime$\downarrow$} & {\small SR$\uparrow$} & {\small RMSE$\downarrow$}   & {\small SR$\uparrow$} & {\small RMSE$\downarrow$}  & {\small SR$\uparrow$} & {\small RMSE$\downarrow$}  & {\small SR$\uparrow$} & {\small RMSE$\downarrow$}     \\ 
 \midrule
 \textbf{SLAM-net (ours)}  &  0.06s  &  \textbf{83.8\pct} & \textbf{0.16m} &  \textbf{62.9\pct} & \textbf{0.28m} & \textbf{77.1\pct} & \textbf{0.19m}  & \textbf{54.3\pct} & \textbf{0.26m} \\ 
 Learned visual odometry    &  0.02s  &  60.0\pct & 0.26m &  24.8\pct & 0.63m &  30.5\pct & 0.47m  & 28.6\pct & 0.40m \\ 
FastSLAM~\cite{montemerlo2003fastslam} & -- & 21.0\pct & 0.58m & 0.0\pct & 3.27m & 21.9\pct & 0.69m & X & X\\
ORB-SLAM~\cite{mur2017orb}  &  0.08s  &  3.8\pct & 1.39m &  0.0\pct & 3.59m & 0.0\pct & 3.54m  & X & X \\ 
Blind baseline            &  0.01s  &  16.2\pct & 0.80m &  1.0\pct & 4.13m & 3.8\pct & 1.50m  & 16.2\pct & 0.80m \\ 

 \bottomrule
\end{tabular}%
}%
{\vspace{0.0cm}\caption{Main SLAM results.}\label{tab:localization}}
\vspace{-0.0cm}
\end{table*}

\begin{table}
\centering
\scalebox{0.9}{
\begin{tabular}{lccc@{\hskip .6cm}c@{\hskip .3cm}c}
\toprule
 & \multicolumn{3}{c@{\hskip .6cm}}{\textbf{Component}} & \textbf{RGBD}  & \textbf{RGB}  \\    
 &  T &  M &  Z & {\small SR$\uparrow$} & {\small SR$\uparrow$}   \\ 
 \midrule
\textbf{SLAM-net (ours)}       & \textbf{l} & \textbf{l} & \textbf{l} &  \textbf{77.1\pct} &  \textbf{55.2\pct}  \\
                               & h & l & l &  43.8\pct &   19.1\pct  \\
                               & l & h & h &  58.1\pct &   X  \\
FastSLAM                       & h & h & h &  21.9\pct &   X  \\

\bottomrule
\end{tabular}%
}%
{\vspace{0.0cm}\caption{Learned vs. handcrafted SLAM components.}\label{tab:fastslam}}
\vspace{-0.0cm}
\end{table}

\begin{table}
\centering
\scalebox{0.9}{
\begin{tabular}{l@{\hskip .6cm}c@{\hskip .6cm}c}
\toprule
 & \textbf{RGBD}  & \textbf{RGB}  \\    
 & {\small SR$\uparrow$} & {\small SR$\uparrow$}   \\ 
 \midrule
Default conditions                  &  3.8\pct & X  \\
No sensor noise                     &  7.5\pct &   18.0\pct   \\
No sensor and actuation noise       &  18.0\pct &  20.4\pct  \\
High frame rate                     &  30.4\pct &  X  \\
Ideal conditions                    &  86.0\pct &   43.5\pct  \\

\bottomrule
\end{tabular}%
}%
{\vspace{0.0cm}\caption{ORB-SLAM results under idealized conditions.}\label{tab:orbslam}}
\vspace{-0.0cm}
\end{table}

\subsection{Baselines}  %

\myparagraph{Learned visual odometry} 
We use the transition model of SLAM-net as a learned visual odometry model.  The model parameterizes a Gaussian mixture that predicts the relative motion between consecutive frames. When the model is used for visual odometry we simply accumulate the mean relative motion predictions. 

\myparagraph{ORB-SLAM}
We adopt the popular ORB-SLAM~\citep{mur2015orb, mur2017orb} as a classic baseline. %
The algorithm takes in RGB or RGB-D images, constructs a keypoint-based sparse map, and estimates a 6-DOF camera pose at every time-step. The algorithm relies on tracking features between consecutive frames. If there are not enough tracked key-points, the system is lost. When this happens we initialize a new map and concatenate it with the previous map based on relative motion estimated from robot actions. With RGB-D input re-initialization takes one time step, with RGB-only input it takes several steps. %

We carefully tuned the hyperparameters of ORB-SLAM based on the implementation and configuration of \citet{mishkin2019benchmarking}, who tuned the algorithm for Habitat simulation data, although without sensor and actuation noise.
For the main localization results in \tabref{tab:localization} we use the default velocity-based motion model as in \citep{mur2015orb}. In \tabref{tab:transfer} we replace the motion model with relative motion estimated from actions, which gave better results.   %

\myparagraph{FastSLAM}
FastSLAM uses the same particle filter algorithm as our SLAM-net, but with handcrafted transition, mapping and observation models. We naively adapt the original algorithm~\citep{montemerlo2003fastslam} to be used with occupancy grid maps. Specifically, we use the same local map representation as in SLAM-net.  The transition model is a Gaussian mixture that matches the ground truth actuation noise. %
The mapping and observation models are then naive implementations of the inverse lidar model and the beam model described in Chapter~9.2 and Chapter~6.3 of \cite{thrun2005probabilistic}, respectively. We create artificial lidar sensor readings by taking the center row of depth inputs. In the observation model instead of pair-wise comparisons we combine the 32 most relevant local maps into a global map. The number of particles are chosen to be the same as for SLAM-net.

\myparagraph{Blind baseline}  
This baseline ignores observation inputs, and instead it accumulates the nominal robot motion based on the ground-truth (but noisy) motion model.
This serves as a calibration of the performance of other methods.

\begin{table*}
\centering
\scalebox{0.9}{
\begin{tabular}{l@{\hskip 1.0cm}cc@{\hskip 1.0cm}cc@{\hskip 1.0cm}cc@{\hskip .8cm}cc}
\toprule
Dataset
  & \multicolumn{2}{c@{\hskip 1.0cm}}{\textbf{Replica}} 
  & \multicolumn{2}{c@{\hskip 1.0cm}}{\textbf{Matterport}} 
  & \multicolumn{2}{c@{\hskip 1.0cm}}{\textbf{Replica}}  
  & \multicolumn{2}{c}{\textbf{Matterport}}  \\    
Sensor
  & \multicolumn{2}{c@{\hskip 1.0cm}}{\textbf{RGBD}} 
  & \multicolumn{2}{c@{\hskip 1.0cm}}{\textbf{RGBD}} 
  & \multicolumn{2}{c@{\hskip 1.0cm}}{\textbf{RGB}}  & \multicolumn{2}{c}{\textbf{RGB}} \\    
Metric  & {\small SR$\uparrow$} & {\small RMSE$\downarrow$}   & {\small SR$\uparrow$} & {\small RMSE$\downarrow$}  & {\small SR$\uparrow$} & {\small RMSE$\downarrow$}  & {\small SR$\uparrow$} & {\small RMSE$\downarrow$}     \\ 
\midrule
 \textbf{SLAM-net (ours)}    &  \textbf{78.8\pct} & \textbf{0.17m} &  \textbf{49.5\pct} & \textbf{0.39m} & \textbf{45.3\pct} & \textbf{0.31m}  & \textbf{23.3\pct} & \textbf{0.54m} \\ 
 Learned visual odometry    &  51.2\pct & 0.31m &  22.4\pct & 0.75m & 17.7\pct & 0.67m  & 15.2\pct & 0.93m \\ 
FastSLAM~\cite{montemerlo2003fastslam}                & 10.0\pct & 0.91m  & 5.7\pct & 1.81m &   X & X  & X & X \\
 ORB-SLAM~\cite{mur2017orb}       &  5.2\pct & 1.46m &  1.9\pct & 2.90m & X & X  & X & X \\ 
Blind baseline            &  7.7\pct & 0.92m &  5.7\pct & 2.3m &  7.7\pct & 0.92m &  5.7\pct & 2.3m \\ 
 \bottomrule
\end{tabular}%
}%
{\vspace{0.0cm}\caption{Transfer results.}\label{tab:transfer}}
\vspace{-0.0cm}
\end{table*}

\section{SLAM results}\label{sec:results}

Main SLAM results are summarized in \tabref{tab:localization}. Visualizations are in the appendix. Videos are available at \url{http://sites.google.com/view/slamnet}.  %

\subsection{Main results for SLAM}

\tabref{tab:localization} reports success rate (SR) that measures the percentage of episodes where the final pose error is below 0.36 meters (to enable successful downstream navigation); and root-mean-square-error (RMSE) which measures the absolute trajectory error as defined in~\citep{handa2014benchmark}. Estimated trajectories are only aligned with the ground-truth at the beginning of each episode. We also report runtimes, measuring the average processing time per frame including loading the data (RGBD sensor, {traj\_expert} data).

\myparagraph{SLAM-net learns successfully} 
We first observe that SLAM-net successfully learned to localize in many episodes despite the challenging data. Comparing columns we see that an imperfect navigation policy can significantly increase the difficulty of localization. Comparing SLAM-net across sensor modalities we find that SLAM-net performs reasonably well with RGB-only input, and the depth sensor helps substantially (54.3\% vs. 83.8\% success for {traj\_expert} data).

\myparagraph{SLAM-net outperforms its alternatives}
We find that SLAM-net outperforms learned visual odometry, the model-based FastSLAM and ORB-SLAM by a large margin across all datasets and sensor modalities; and its runtime (on GPU) is slightly better than ORB-SLAM. Next we discuss the comparison with FastSLAM and ORB-SLAM in detail.

\subsection{Learned vs. handcrafted SLAM components}
Interestingly SLAM-net outperforms FastSLAM. The algorithm is exactly the same, the difference is that FastSLAM has simple handcrafted model components, while SLAM-net has neural network components that are learned end-to-end.

\tabref{tab:fastslam} combines learned (l) and handcrafted (h) alternatives  for each of the model components: transition model (T), mapping model (M), observation model (Z). SLAM-net has all models learned, FastSLAM has all models handcrafted. We report results for the {traj\_nav} data. We find that learning any of the model components is useful, and learning all model components jointly gives the best performance. This can be attributed to both model representation and task-oriented learning. First, our neural networks may encode a more powerful function than our handcrafted models. Second, our we learn models end-to-end, so they are optimized for the task in the context of the algorithm and the dataset. For RGB-only input we do not have handcrafted mapping and observation models, but SLAM-net is able to learn effective models end-to-end.

\subsection{Why does ORB-SLAM fail?}

ORB-SLAM relies on temporal coherence between frames to track features, which is challenging in our domain due to the combined effects of sensor noise,  sparse visual features, rapid turns (approx. $90^{\circ}/s$), low frame rate (approx. 3 fps), and narrow field of view (HFOV=$70^\circ$). We find that ORB-SLAM often fails to track features even with RGB-D input. With RGB-only input it fails in nearly all steps, hence we could not report a meaningful result. %
In contrast to ORB-SLAM, SLAM-net does not rely explicitly on feature tracking, and it learns task-oriented models that can, \eg,  learn more robust feature extractors for this domain.

In \tabref{tab:orbslam} we evaluate ORB-SLAM in idealized settings for the {traj\_expert} data. Each row removes different types of challenges: no sensor noise, no actuation noise, high frame rate, and ideal condition. The high frame rate setting reduces the action step size in Habitat to achieve an equivalent 3 to ${\sim}30$ fps increase in frame rate. The ideal condition setting removes all the above challenges together. Our results show that ORB-SLAM only works well in ideal conditions, where its performance is comparable to SLAM-net in the hardest conditions. %
If we remove only one type of challenge the performance remains significantly worse. The RGB-D results indicate that low frame rate has the largest impact. %
For RGB the trend is similar, but the presence of observation noise makes feature tracking fail completely. %

\subsection{Transfer results}\label{sec:transfer}

An important concern with learning based SLAM method is potential overfitting to the training environments. We take the SLAM-net models trained with the Gibson data, and evaluate them for the Replica and Matterport datasets, with no additional training or hyperparameter tuning. These datasets contain higher quality scenes and cover a wide range of smaller (Replica) and larger (Matterport) indoor environments. The robot and camera parameters remain the same.

Results are in \tabref{tab:transfer}. We observe strong performance for SLAM-net across all datasets and sensor modalities. Comparing alternative methods we observe a similar trend as for the Gibson data ({traj\_expert} columns in \tabref{tab:localization}). Note that results across datasets are not directly comparable as the length of trajectories differ, \eg, they are longer in the Matterport data (see \tabref{tab:datastat} for statistics). We believe that these results on photorealistic data are promising for sim-to-real transfer to real robot navigation data.

\subsection{Ablation study}

To better understand the workings of SLAM-net we perform a detailed ablation study. Results are summarized in \tabref{tab:ablation}. The table reports success rates for the Gibson {traj\_nav} data, using SLAM-net in different conditions.

\myparagraph{Joint training is useful} 
Line (2) of \tabref{tab:ablation} evaluates the pre-trained transition and observation models without joint finetuning. We find that finetuning is useful, and its benefit is much more significant for RGB input. A possible explanation is that our RGB model uses maps with latent features, while the RGBD model uses both latent features and predicted occupancy. Without finetuning, the occupancy predictor may generalize better to our evaluation setting, where the overall uncertainty is higher than during pre-training.

\myparagraph{Occupancy maps are useful for localization}
Lines (3--5) use different channels in learned local maps, pre-trained occupancy predictions, learned latent features, or both. The RGBD model is comparable in all settings. Adding latent maps on top of occupancy only improves 1.9\%, which indicates that 2D occupancy is sufficient for localization. The latent map configuration is $4.7\%$ behind the occupancy maps, showing that we can learn useful map features end-to-end without direct supervision.  

\myparagraph{We can learn better map features if occupancy prediction is difficult}
Comparing the RGB models we find that the occupancy maps do not perform well here, but end-to-end training allowed learning more effective features. The difference to RGBD can be explained by the substantially lower prediction accuracy of our occupancy map predictions.

\myparagraph{Choosing what to compare matters}
Lines (6--9) compare strategies for choosing which map-pose pairs to feed into our discriminative observation model. Line (6) uses the last 8 steps of the particle trajectory. Lines (7--9) choose the most relevant past steps based on their estimated overlapping view area. As expected, dynamically choosing what to compare is useful. While one would expect more comparisons to be useful, over 8 comparisons do not improve performance. Since we trained with 8 comparisons, this result indicates that our model overfits to this training condition.

\myparagraph{More particles at inference time are useful}
Lines (10--16) vary the number of particles at inference time. Surprisingly, as little as 8 particles can already improve over the visual odometry setting (line 10). Increasing the number of particles helps, providing a trade-off between performance and computation. The effect for RGB is less pronounced,  and improvement stops over 128 particles.

\begin{table}
\centering
\scalebox{0.9}{
\begin{tabular}{ll@{\hskip .6cm}c@{\hskip .6cm}c}
\toprule
\multicolumn{2}{l}{Sensor} & \textbf{RGBD}  & \textbf{RGB}  \\    
\multicolumn{2}{l}{Metric}  & {\small SR$\uparrow$} & {\small SR$\uparrow$}   \\ 
 \midrule
 (1) & SLAM-net (default)         &  \textbf{77.1\pct} &  \textbf{55.2\pct} \\
 (2) & No joint training            &  66.7\pct &   8.6\pct   \\

\smallgap
(3) & Occupancy map only          &  75.2\pct &   23.8\pct  \\
(4) & Latent map only             &  70.5\pct &   \textbf{55.2\pct}  \\
(5) & Occupancy + latent map      &  \textbf{77.1\pct} &   44.8\pct    \\
\smallgap
(6) & Fixed comparisons (8)     &  44.8\pct  &   29.5\pct   \\
(7) & Dynamic comparisons (4)   &  73.3\pct  &  41.9\pct  \\
(8) & Dynamic comparisons (8)   &  \textbf{77.1\pct}  &  \textbf{55.2\pct}  \\
(9) & Dynamic comparisons (16)  &  \textbf{77.1\pct}  &  40.0\pct  \\
\smallgap
(10) & K=1 (VO)         &  30.5\pct    &  26.7\pct   \\
(11) & K=8              &  60.0\pct    &  35.2\pct  \\
(12) & K=32 (training)  &  72.4\pct    &  39.1\pct  \\
(13) & K=64             &  75.2\pct    &  46.7\pct  \\
(14) & K=128 (evaluation default)  &  77.1\pct    &  \textbf{55.2\pct}  \\
(15) & K=256            &  79.1\pct    &  44.8\pct  \\
(16) & K=512            &  \textbf{82.9\pct}    &  48.6\pct  \\
\bottomrule
\end{tabular}%
}%
{\vspace{0.0cm}\caption{Ablation results.
}\label{tab:ablation}}
\vspace{-0.0cm}
\end{table}

\subsection{KITTI odometry results}
To better understand the limitations of our approach we apply it to the KITTI odometry data, which contains long trajectories of autonomous driving. We do not expect a strong performance. SLAM-net is designed to enable indoor robot navigation which is reflected in a number of design decisions. First, our local maps ignore information far from the camera. Second, we do not have a dedicated particle proposal mechanism for closing large loops. Third, a key benefit of our approach is the joint training of its components, however, this requires large and diverse training data. The KITTI data is relatively small for this purpose. Finally, images in the KITTI data are of high quality for which existing SLAM methods are expected to work well. 

Our results are in \tabref{tab:kitti}. We report RMSE in meters after trajectory alignment. SLAM-net results are averaged over 5 seeds. The ORB-SLAM results are for RGB only input taken from~\cite{mur2015orb}.
As expected,  SLAM-net does not perform as well as ORB-SLAM; nevertheless, it learns a meaningful model and outperforms learned visual odometry. %
Looking at predicted trajectories we find that SLAM-net occasionally fails to capture turns of the road (visualizations are in the Appendix). 
One reason is that there are no particles near the true trajectory, or the observation model gives a poor prediction.  The training data contains only a limited number of turns, which makes learning from scratch difficult. Indeed our model starts to overfit after a few epochs, suggesting that more training data would improve the performance.

\begin{table}
\centering
\scalebox{0.9}{
\begin{tabular}{l@{\hskip .6cm}cc}
\toprule
Trajectory  & \textbf{Kitti-09} & \textbf{Kitti-10}  \\
Metric  & {\small RMSE$\downarrow$} & {\small RMSE$\downarrow$}  \\ 
\midrule
\textbf{SLAM-net (ours)}                 &  83.5m     & 15.8m   \\
SLAM-net (best of 5)                     &  56.9m     & 12.8m  \\
  Learned visual odometry                  &  71.1m      &  73.2m  \\
ORB-SLAM-RGB~\cite{mur2015orb}           &  \textbf{7.62m}       & \textbf{8.68m}    \\
\bottomrule
\end{tabular}%
}%
{\vspace{0.0cm}\caption{KITTI results.}\label{tab:kitti}}
\vspace{-0.0cm}
\end{table}

\section{Navigation results}\label{sec:navresults}
The motivation of our work is to enable visual robot navigation in challenging realistic conditions. Our navigation results are reported in \tabref{tab:navigation} and \tabref{tab:leaderboard}, using RGB-D input. Videos are on the project website.
We report two key metrics following~\citet{anderson2018evaluation}: success rates (SR) and success weighted path length (SPL). 

In \tabref{tab:navigation} we experiment with our navigation pipeline using  different methods for localization and mapping, but keeping the planner and controller modules fixed. Navigation performance is strong with a ground-truth localization oracle, which validates our architecture and serves as an upper-bound for SLAM methods. The navigation architecture with SLAM-net significantly outperforms visual odometry, achieving 65.7\% success.  Our navigation architecture with visual odometry is conceptually similar to that of Active Neural SLAM~\cite{chaplot2020learning} and Occupancy Anticipation~\cite{ramakrishnan2020occupancy}. %
Our results are consistent with that of \citet{ramakrishnan2020occupancy} in matching conditions. 
We did not compare with the classic ORB-SLAM method here because of its poor performance in our previous experiments.

Finally, we submitted our method to the Habitat Challenge 2020 evaluation server, which allows direct comparison with various alternative methods. \tabref{tab:leaderboard} shows the top of the leaderboard for the PointNav task. SLAM-net achieves $64.5\pct$ success, significantly improving over the SOTA (VO~\cite{vomethod}, $37.3\pct$). It also outperforms the challenge winner (OccupancyAnticipation~\cite{ramakrishnan2020occupancy}, $29.0\pct$) which was shown to be superior to Active Neural SLAM~\cite{chaplot2020learning}.

\begin{table}[tb]%
\centering
\scalebox{0.9}{
\begin{tabular}{lcc}
\toprule
 \textbf{SLAM component}  & {\small SR$\uparrow$} & {\small SPL$\uparrow$}  \\ 
 \midrule
Ground-truth   &  \textbf{90.7\pct} &  \textbf{0.56} \\
\textbf{SLAM-net (ours)}   &  \textbf{65.7\pct} &  \textbf{0.38} \\
Learned visual odometry   &  32.4\pct &  0.19 \\
 \bottomrule
\end{tabular}
}{\vspace{0.0cm}\caption{Navigation results.}\label{tab:navigation}}%
\end{table}

\begin{table}[tb]%
\centering
\scalebox{0.9}{
\begin{tabular}{llcc}
\toprule
 \textbf{Rank} & \textbf{Method}  & {\small SR$\uparrow$} & {\small SPL$\uparrow$}  \\ 
 \midrule
1 & \textbf{SLAM-net (ours)}                &  \textbf{64.5\pct} &  \textbf{0.377} \\
2 & VO~\cite{vomethod}                                           &  37.3\pct &  0.266 \\
3 & OccupancyAnticipation~\cite{ramakrishnan2020occupancy}      &   29.0\pct &  0.220 \\
\bottomrule
\end{tabular}
}{\vspace{0.0cm}\caption{Habitat 2020 PointNav Challenge leaderboard accessed on 16 November 2020~\cite{habitat2020leaderboard}.}\label{tab:leaderboard}}%
\end{table}

\section{Conclusions}
We introduced a learning-based differentiable SLAM approach with strong performance on challenging visual localization data and on downstream robot navigation, achieving SOTA in the Habitat 2020 PointNav task.

Together, our results provide new insights for understanding the strengths of classic and learning based SLAM approaches in the context of visual navigation.
Our findings %
partially contradict the results of \citet{mishkin2019benchmarking}, who benchmarked classic and learned SLAM for navigation. While they found ORB-SLAM to be better than learning based SLAM in the same Habitat simulator, they used a noise-free setting and relative goals. As pointed out by \citet{habitatchallenge}, this setting is not realistic. Indeed, we tried running the public ORB-SLAM implementation of \citet{mishkin2019benchmarking} in our simulator setting and it failed completely; while our learning-based approach achieved strong performance. 

We believe that our work on differentiable SLAM may lay foundation to a new class of methods that learn robust, task-oriented features for SLAM. Future research may investigate alternative differentiable SLAM algorithms, \eg, that build on an optimization-based method instead of particle filtering. 
While our initial results are promising, future work is needed to apply SLAM-net to real-world robot navigation. A particularly interesting application would be learning to relocalize with significant changes in the environment, a setting known to be challenging for existing SLAM algorithms.

\subsection*{Acknowledgement}
{\small
We would like to thank Rico Jonschkowski for suggesting to keep local maps and Gim Hee Lee for valuable feedback. This research/project is supported in part by the National Research Foundation, Singapore under its AI Singapore Program (AISG Award No: AISG2-RP-2020-016) and by the National University of Singapore (AcRF grant R-252-000-A87-114).
}

{\small
\bibliographystyle{plainnat}
\bibliography{slamnet}
}

\clearpage
\onecolumn

\begin{appendices}

\section{SLAM-net components}

\subsection{Observation model}
In the particle filter the observation model estimates the log-likelihood $\log{w_t^k}$, the probability of the current observation $o_t$ given a particle trajectory $s_{1:t}^k$ and past observations $o_{1:{t-1}}$. 
The particle weight is multiplied with the estimated log-likelihood. 

In SLAM-net we decompose this function. A mapping model first predicts a local map $m_t$ from $o_t$. The observation model than estimates the compatibility of $m_t$ with $m_{1:t-1}$ and $s_{1:t}^k$, by summing pair-wise compatibility estimates. 
\begin{equation}
    \log{w_t^k} \approx \sum_{\tau \in T}{\log{w_{t, \tau}^k}}
\end{equation}
\begin{equation}
    \log{w_{t, \tau}^k} = f^{\mathrm{obs}}_\theta(m_t, s_t^k, m_{t-\tau}, s_{t-\tau}^k)
\end{equation}

The network architecture of  $f^{\mathrm{obs}}_\theta$ is shown in~\figref{fig:networks}. Weights are shared across particles and time steps. The important component of this model is the image transformation (Transform), that applies translational and rotations image transformations to $m_{t-\tau}$ given the relative poses defined by $s_t^k$ and $s_{t-\tau}^k$. We use spatial transformer networks~\cite{jaderberg2015spatial} that implement these transforms in a differentiable manner.

\subsection{Mapping model}
The mapping model component takes in the $o_t$ observation ($160\mytimes 90$ RGB or depth image) and outputs a local map $m_t$. Local maps are $40\mytimes40\mytimes \mapchannels$ grids, which can be understood as images with $\mapchannels$ channels. The local maps either encode latent features ($\mapchannels=16$) or they are trained to predict the occupied and visible area. 
Each cell in the local map corresponds to a $12\mytimes12$cm area in front of the robot, \ie, the local map covers a $4.8\mytimes4.8$m area. 

The network architecture (configured for latent local maps) is shown in \figref{fig:networks}. The same network architecture is used for RGB and depth input. We apply a fixed perspective transform to the first-person input images, transforming them into $160\mytimes 160$ top-down views that cover a $4.8\mytimes 4.8$m area. The perspective transformation assumes that the camera pitch and roll, as well as the camera matrix are known.

When SLAM-net is configured with local maps that predict occupancy we use a similar architecture but with separately learned weights. In case of RGB input our network architecture is the same as the mapping component in ANS~\cite{chaplot2020learning}, using the same ResNet-18 conv-dense-deconv architecture. We freeze the first three convolutional layer of the ResNet to reduce overfitting. In case of depth input our network architecture is similar to the one in \figref{fig:networks}, but it combines  top-down image features with first-person image features. When SLAM-net is configured with both occupancy and latent local maps we use separate network components and concatenate the local map predictions along their last (channel) dimension.

For the KITTI experiments we use  $40\mytimes40$ local maps where cells are $70\mytimes70$cm, i.e., a local map captures a $28\mytimes28$m area. We use an equivalent network architecture that is adapted to the wider input images.

\subsection{Transition model}
The transition model takes in the last two consecutive observations ($o_t$, $o_{t-1}$) and it outputs a distribution over the relative motion components of the robot.  It can also take in the last robot action $a_{t-1}$ when it is available. 

\begin{figure*}[ht]
\centering
\vspace{0.5cm}
\includegraphics[width=0.98\textwidth]{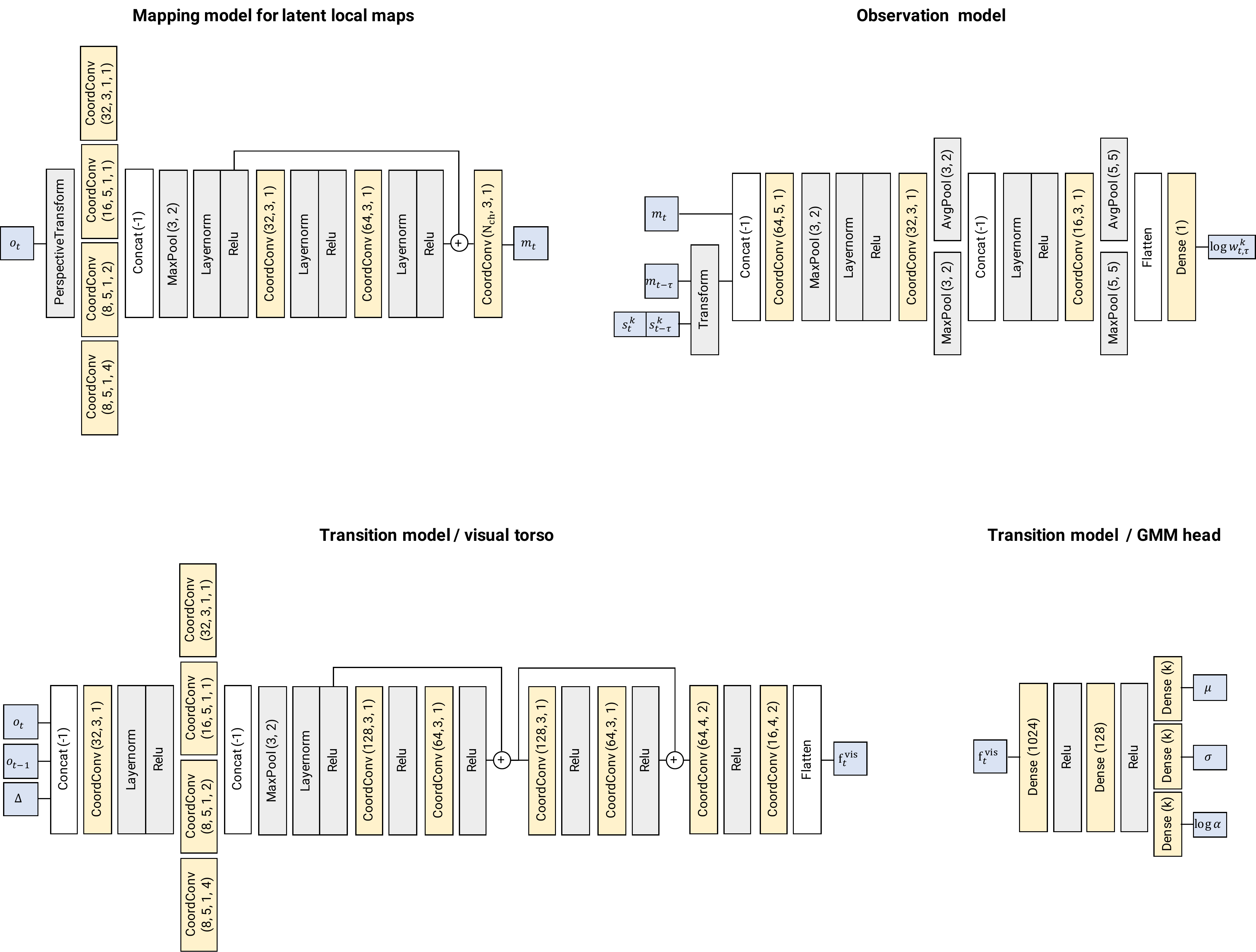}
\vspace{0.5cm}
\caption{\textbf{Network architectures of SLAM-net components.} The notation is as follows. \textbf{Conv}(f, k, s, d): convolutional layer with f filters, k$\mytimes$k kernel, s$\mytimes$s strides, d$\mytimes$d dilation. \textbf{CoordConv}: convolutional layer that takes in extra channels that encode constant pixel coordinates. \textbf{MaxPool}(k, s): max-pooling layer with k$\mytimes$k kernel and s$\mytimes$s strides.  \textbf{AvgPool}(k, s): average-pooling layer with k$\mytimes$k kernel and s$\mytimes$s strides. \textbf{Dense}(h) dense fully-connected layer with h output units. \textbf{Concat}(-1): concatenate input images along their channel axis. \textbf{Transform}: spatial image transformation using a spatial transformer network~\cite{jaderberg2015spatial}. Given a relative pose and orientation it applies translational and rotational image transformations. \textbf{PerspectiveTransform}: geometric perspective transformation to a top-down view, given a known camara matrix and a target resolution. \textbf{Inputs and outputs}. $o_t$: input image at time $t$, either RGB or depth. $m_t$: local map at time $t$. $s_t^k$: state $s$ along particle trajectory for particle $k$ and time $t$. $\log{w_{t,\tau}^k}$: estimated compatibility of map-pose pairs at time $t$ and $t-\tau$ for particle $k$.  $\Delta$: difference of input image pair $\dense o_t - o_{t-1}$. $\mathrm{f^{vis}}_t$: intermediate visual features in the transition model. $\mu$: mean predictions for a Gaussian mixture model (GMM) with $\dense k=3$ components. $\sigma$: predicted standard deviations. $\mathrm{log}\alpha$: mixture log-probabilities.
}
\label{fig:networks}
\end{figure*}

Our transition model parameterizes a Gaussian mixture model (GMM) with $\dense k=3$ mixture components, separately for each coordinate (x, y, yaw) and each discrete robot action $a$. The network architecture is shown in \figref{fig:networks}. The model first extracts visual features $\mathrm{f^{vis}}_t$ from a concatenation of $o_t$, $o_{t-1}$ and their difference $\dense \Delta=o_t - o_{t-1}$. The visual features are then fed to GMM heads for each combination of robot action $a$ and x, y, yaw motion coordinates. Each GMM head uses the same architecture with independent weights.

\clearpage
\section{Additional figures}

\begin{figure*}[h]
\centering
\includegraphics[width=0.98\textwidth]{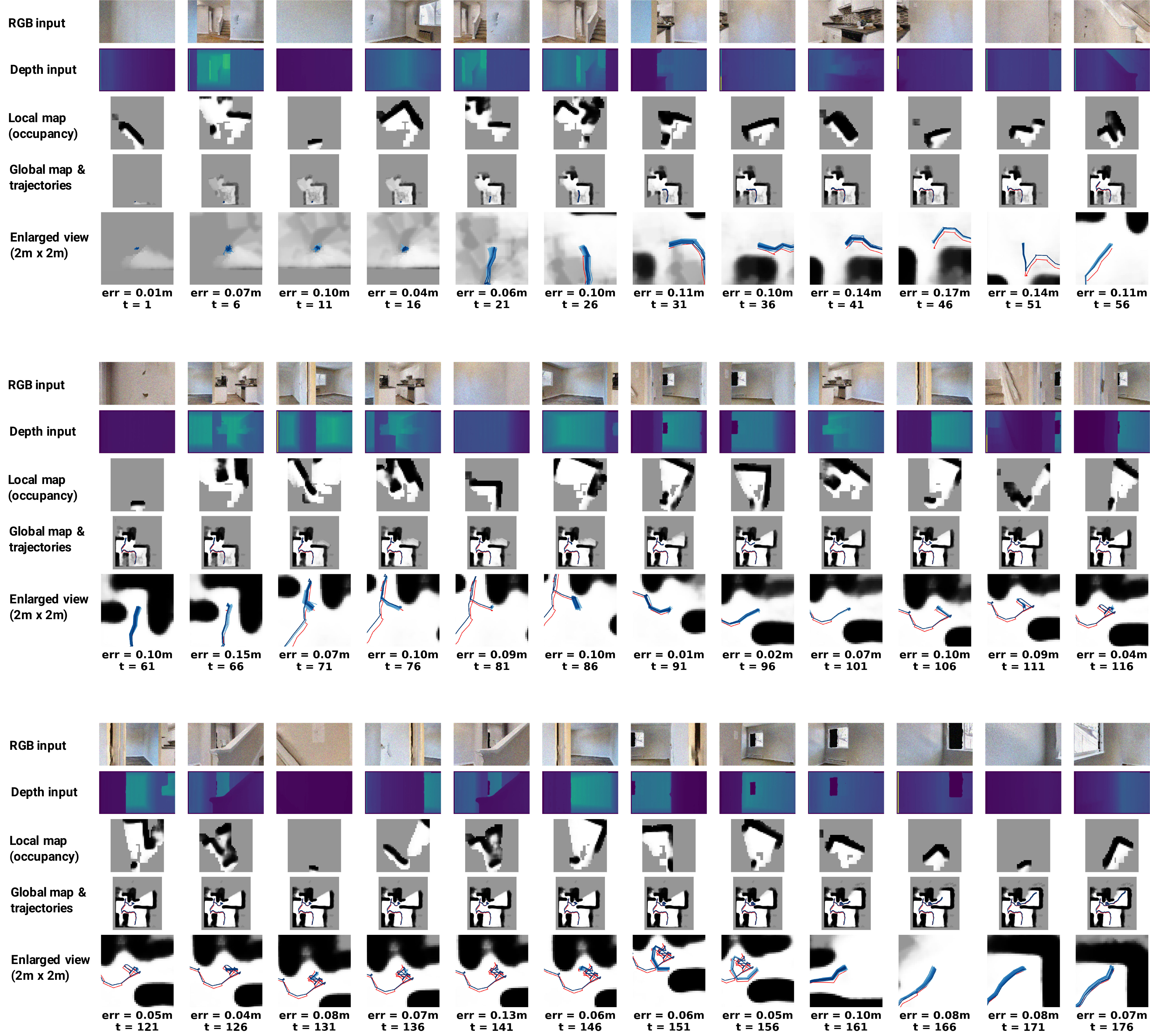}
\caption{\textbf{SLAM-net with RGBD sensor.} Trajectory sample from the Gibson data traj\_exp\_rand test set. In this configuration SLAM-net local maps predict occupancy (third row). The enlarged view (bottom row) shows a $2\mytimes 2$m window of the predicted global map. The true trajectory is in red; particle trajectories are in blue. The shade of blue indicate particle weights (darker color means larger particle weight).
 Notice that particle trajectories maintain multi-modal trajectories. Low weight particles are dropped after resampling.  The figure is best viewed using zoom.
}
\label{fig:example_depth}
\end{figure*}

\begin{figure*}[h]
\centering
\includegraphics[width=0.98\textwidth]{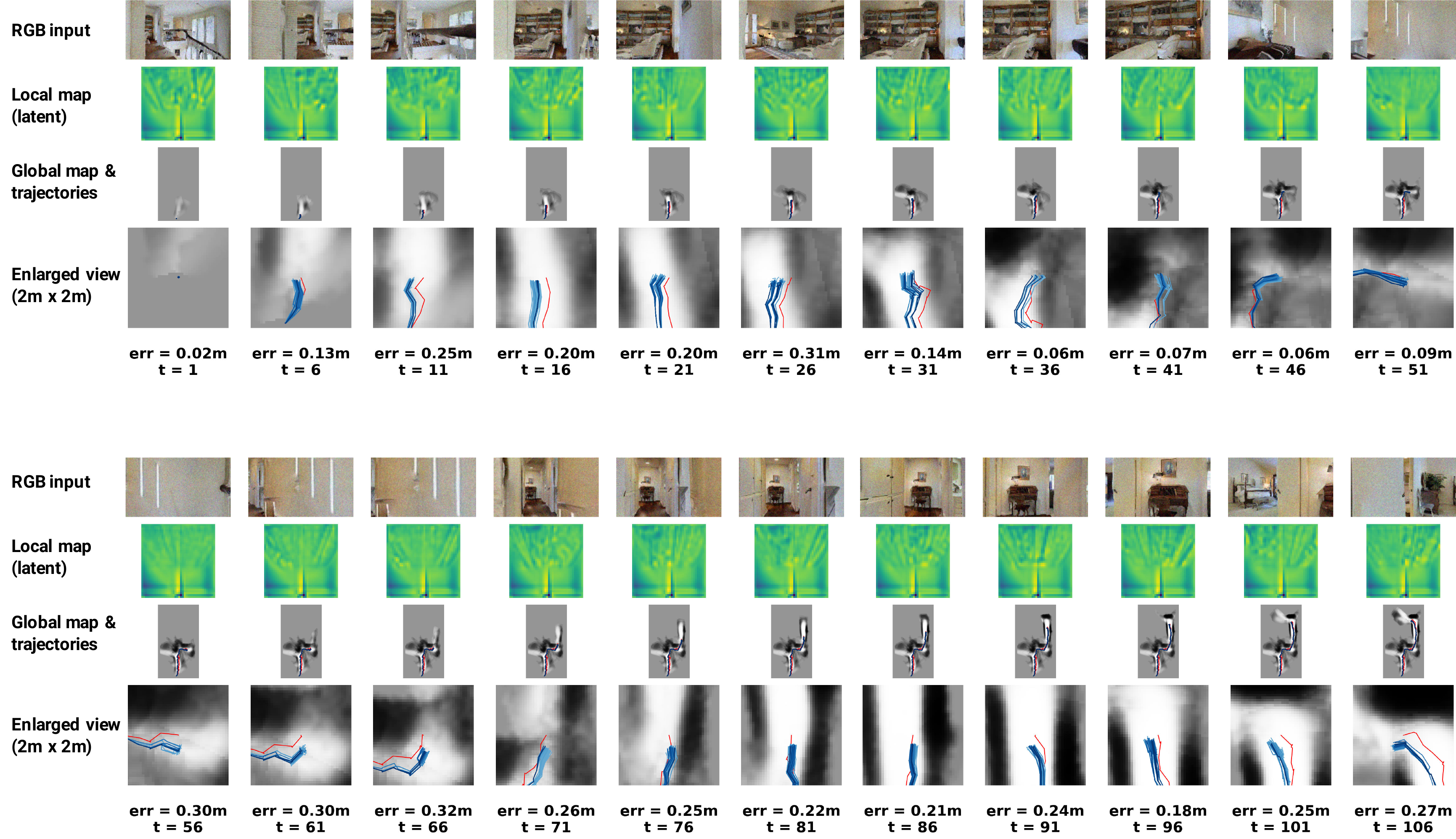}
\caption{\textbf{SLAM-net with RGB only sensor.} Trajectory sample from the Gibson data traj\_exp test set. In this configuration SLAM-net local maps have 16 latent feature channels. We visualize normalized features of a single channel (second row). The enlarged view (bottom row) shows a $2\mytimes 2$m window of the predicted global map. The true trajectory is in red; particle trajectories are in blue. The shade of blue indicate particle weights (darker color means larger particle weight). Notice that particle trajectories maintain multi-modal trajectories. Low weight particles are dropped after resampling. The figure is best viewed using zoom.
}
\label{fig:example_rgb}
\end{figure*}

\begin{figure*}[h]
\centering
\includegraphics[width=0.8\textwidth]{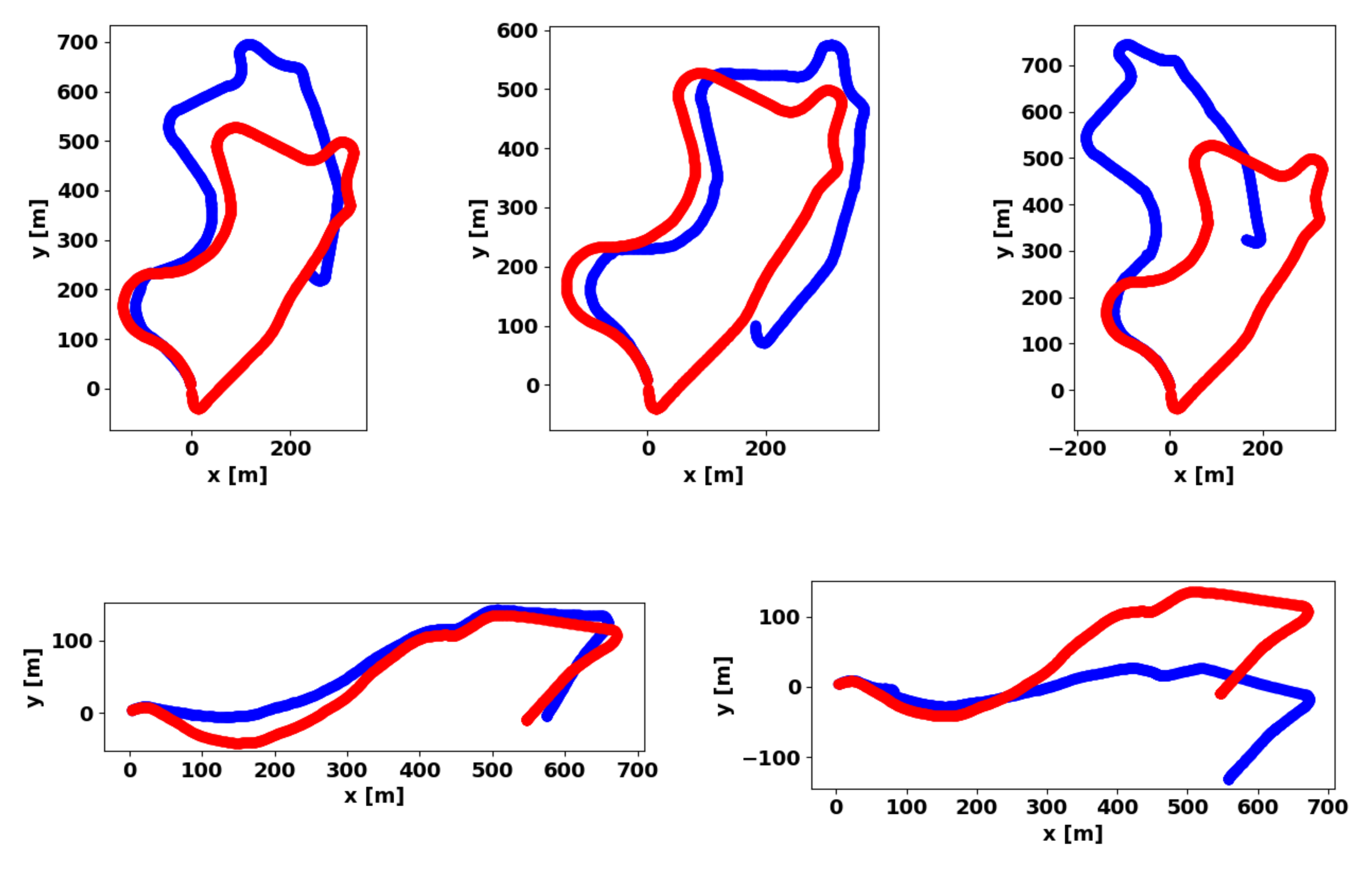}
\caption{\textbf{KITTI trajectories.} 
The figure shows SLAM-net predictions (before alignment) with different random seeds for the Kitti-09 trajectory (top row) and the Kitti-10 trajectory (bottom row). The predicted trajectory is in blue, the true trajectory is in red. }
\label{fig:kitti_examples}
\end{figure*}

\end{appendices}

\end{document}